\documentclass[manuscript]{acmart}

\usepackage{graphicx}
\usepackage{tabularx}

\copyrightyear{2023}
\acmYear{2023}
\setcopyright{rightsretained}
\acmConference[UMAP '23 Adjunct]{Adjunct Proceedings of the 31st ACM Conference on User Modeling, Adaptation and Personalization}{June 26--29, 2023}{Limassol, Cyprus}
\acmBooktitle{Adjunct Proceedings of the 31st ACM Conference on User Modeling, Adaptation and Personalization (UMAP '23 Adjunct), June 26--29, 2023, Limassol, Cyprus}\acmDOI{10.1145/3563359.3597403}
\acmISBN{978-1-4503-9891-6/23/06}

\begin{document}

\title[Development of a Trust-Aware User Simulator]{Development of a Trust-Aware User Simulator for Statistical Proactive Dialog Modeling in Human-AI Teams (Preprint)}

\author{Matthias Kraus}
\affiliation{%
  \institution{ University of Augsburg}
  \city{Augsburg}
  \country{Germany}
  }
\email{matthias.kraus@uni-a.de}

\author{Ron Riekenbrauck}
\affiliation{%
  \institution{Ulm University}
  \city{Ulm}
  \country{Germany}
  }
\email{ron.riekenbrauck@uni-ulm.de}

\author{Wolfgang Minker}
\affiliation{%
  \institution{Ulm University}
  \city{Ulm}
  \country{Germany}
  }
\email{wolfgang.minker@uni-ulm.de}

\renewcommand{\shortauthors}{Kraus et al.}

\begin{abstract}
  The concept of a Human-AI team has gained increasing attention in recent years. For effective collaboration between humans and AI teammates, proactivity is crucial for close coordination and effective communication. However, the design of adequate proactivity for AI-based systems to support humans is still an open question and a challenging topic. In this paper, we present the development of a corpus-based user simulator for training and testing proactive dialog policies. The simulator incorporates informed knowledge about proactive dialog and its effect on user trust and simulates user behavior and personal information, including socio-demographic features and personality traits. Two different simulation approaches were compared, and a task-step-based approach yielded better overall results due to enhanced modeling of sequential dependencies. This research presents a promising avenue for exploring and evaluating appropriate proactive strategies in a dialog game setting for improving Human-AI teams.
\end{abstract}





\maketitle

\section{Introduction}
The concept of a Human-AI team (HAIT) is intriguing, but despite the availability of sophisticated conversational assistants such as Alexa, Siri, and ChatGPT, we are still far from achieving an AI that is not only able to solve specific tasks but could also socialize and form a personal bond with its user to build an effective team. HAIT requires close coordination between humans and AI teammates to work together towards a common goal \cite{zhang2021ideal}. Effective communication, prediction of teammates' actions, and high-level coordination are essential components of this collaborative effort. In this regard, the proactive behavior of AI-based systems and the communication thereof during collaboration is an important research topic concerning HAITs, e.g., see Horvitz et al. \cite{horvitz1999principles}. Proactivity can be defined as an AI's self-initiating, anticipatory behavior for contributing to effective and efficient task completion. It has been shown to be essential for human teamwork as it leads to higher job and team performance and is associated with leadership and innovation~\cite{crant2000proactive}. However, the design of adequate proactivity for AI-based systems to support humans is still an open question and a challenging topic. It is essential to study the impact of proactive system actions on the human-agent trust relationship and how to use information about an AI agent's perceived trustworthiness to model appropriate proactive dialog strategies for forming effective HAITs.
There are different experimental approaches to achieving this goal, such as developing strategies and testing them with real users or training statistical-based strategies and testing them using simulated experimental approaches \cite{mctear2020conversational}. Experiments with real users can either take place under laboratory conditions or "in the wild," i.e., in real-life scenarios. However, there are disadvantages to both approaches, including the lack of real-life usage in laboratory conditions and the difficulty of obtaining a full-functioning system during the early stages of development for in-the-wild studies. In addition, recruiting a sufficient number of users to allow a valid interpretation of data is a challenge. To overcome this challenge, user simulation techniques have been developed \cite{eckert1997user,levin2000stochastic,schatzmann2006survey,kreyssig2018neural, jain2018user}. These techniques allow the testing of dialog agent prototypes with a large number of simulated "subjects" and facilitate the exploration of dialog strategies that may enhance HAIT.

In this paper, we present the development of a corpus-based user simulator for training and testing proactive dialog policies. To create the simulator, we collected a proactive dialog corpus with user trust annotations \cite{kraus2022prodial}, utilizing informed knowledge about proactive dialog and its effect on user trust from previous studies \cite{kraus2020effects,kraus2021role,kraus2022kurt,kraus2020context} to ensure high-quality data collection. Our main goal was to replicate realistic user characteristics, tasks, and trusting behaviors for exploring and evaluating appropriate strategies in a dialog game setting that we designed as a sequential decision-making task in a company management environment \cite{kraus2022prodial}. We simulated user behavior and personal information, including socio-demographic features and personality traits, using relevant data from the corpus collection. This enabled us to estimate the current trustworthiness of system behavior \cite{kraus2021modelling} and integrate trust in the dialog state and reward function for creating trust-adaptive proactive dialog strategies \cite{kraus2022improving}. To simulate the user's task behavior, we developed and compared two different simulation approaches: A complexity-based method and a task-step-based method. Both methods were found to be applicable for training statistical proactive dialog strategies, but the task-step-based approach yielded better overall results due to better modeling of sequential dependencies.
\section{Related work}
Over the past 25 years, user simulation for training statistical and primarily RL-based dialog systems has been extensively studied \cite{singh2002optimizing}. The primary focus has been on task-oriented dialog systems that are designed to assist users in achieving a specific task or goal. Examples of such systems include conversational agents used for hotel room bookings or ordering food from a restaurant \cite{young2013pomdp}. These systems conduct so-called slot-filling dialogs, where the system retrieves specific values for pre-defined entities, or slots, of a particular domain, such as food type, price range, and location, via dialog with the user for providing the desired information.
In task-oriented dialog systems, user utterances are typically encoded in semantic representations by a natural language understanding module. The dialog management module then takes these representations as input for filling the respective semantic slots and selecting an appropriate system response according to a specific dialog policy. Therefore, user simulation for training and evaluating task-oriented systems usually produces output in the form of semantic representations of user actions \cite{eckert1997user,levin2000stochastic,lee2012unsupervised, lin2021domain}. However, there are also alternative approaches that can generate natural language utterances \cite{kreyssig2018neural, shi2019build}. \cite{lopez2003assessment} even presented a user simulator that generates spoken utterances based on pre-recorded speech files. User simulators can be classified into two types: rule-based and corpora-based \cite{schatzmann2006survey}. Rule-based methods rely on hand-crafted rules and a range of user profiles \cite{pietquin2005framework}, while corpora-based approaches use probabilistic data-driven methods to model natural user behavior \cite{schatzmann2006survey,lee2012unsupervised}. More recent approaches use sequence-to-sequence or transformer models to produce output on a semantic or natural language level \cite{kreyssig2018neural,shi2019build,lin2021domain}.
Social behavior modeling is also of interest when it comes to user simulation. \citet{egges2004generic} simulated personality, mood, and emotion, and \citet{ferreira2015reinforcement} incorporated social signals in the reward function of the RL-based system for training socially-effective dialog behavior. Several social signals, such as user satisfaction, rapport-building, small talk, and self-disclosure, have been used in different user simulation approaches \cite{ultes2015quality, sun2021simulating, jain2018user}.
We also aim to build a user simulator that can express task-related behavior and model specific user characteristics. However, as our interaction with the agent only involves a limited set of user actions, we opted for a simplistic combination of rule-based and data-driven mechanisms for simulating adequate user behavior. We compared two simulation approaches: using the complexity-based method, the user's task behavior was simulated depending on the agent's action and the complexity of a task step, while the task-step-based method incorporated information from a particular task step and the agent's action. For deciding, which approach to use for training our RL-based proactive dialog agent, we evaluated the approaches according to common metrics for estimating the quality of user simulators \cite{pietquin2013survey}. Here, we utilized the mean square error (MSE) and Kullback-Leibler (KL) Divergence for measuring the error and difference between the behavior of real users and simulated users.

\section{Simulated Proactive Dialog Environment}
As we chose to utilize a corpus-based method to appropriately model a user's task and trusting behavior, it was necessary to have access to a proactive dialog corpus that includes trust annotations. Although there are various data corpora that exist for conventional dialog modeling \cite{williams2014dialog,budzianowski2018multiwoz}, none of them are sufficient for modeling proactive dialog. This is due to the fact that proactive behavior is either absent or underrepresented in these corpora \cite{balaraman2020proactive}, and trust-related features are not adequately annotated. To fill this gap, a new data corpus was created in previous work \cite{kraus2022prodial}, which involved developing an AI agent prototype for personal advising with a proactive dialog model. The agent collected personal and dialog data in a serious gaming scenario, resulting in a trust-annotated data corpus containing interactions with the proactive assistance system. To accurately model and predict trust during mixed-initiative dialog, a trust prediction model was required to simulate a user's trusting behavior. Previous research on trust has identified various user-, system-, and context-related factors that influence the trust relationship \cite{parasuraman1997humans, muir1994trust, lee2004trust, hoff2015trust}. For including this in the development of proactive dialog policies, we collected the necessary features during data collection and build a trust estimation model in previous work \cite{kraus2021modelling}.
The next sections provide further details regarding the data collection method, the corpus, and the techniques employed to predict the HCT relationship.
\subsection{Data Collection}
We acquired a dataset of 308 dialogs, which included a total of 3696 exchanges between users and a proactive dialog agent \cite{kraus2022prodial}. The data was gathered using an online game that utilized the clickworker framework~\footnote{www.clickworker.de}, where users had to make strategic decisions to manage a company with the agent's assistance. Each exchange was marked with the user's self-reported measures of trust in the system, including competence, predictability, and reliability. The data also included objective features such as task complexity, exchange duration, user actions, and static user information such as age, gender, personality, and domain expertise. The game was designed as a turn-based planning task where the user had to make decisions based on the agent's provided options. The agent selected various proactive dialog action types and employed natural language to offer suggestions and information to help the user make the best choice. We made sure that the agent's suggestions generated the most points based on the task step to avoid any unintended negative impacts on the system's trustworthiness.
\subsection{Trust Modelling and Estimation}
Based on the collected data set, we created a new user model to predict online trust when interacting with proactive AI agents \cite{kraus2021modelling}. For this, we selected corpus parameters as features for prediction, including both numerical values (such as age and trust propensity) and categorical values (such as proactivity type). The resulting feature vector contained 57 features, which were categorized as personal user parameters, interaction parameters, and temporal interaction parameters. To predict trust, the variables for trust, competence, reliability, and predictability were combined to create a target variable with a range of 1 to 5 on a Likert-scale. The prediction problem was treated as a multi-class classification task with distinct trust values as target classes. We applied a Support Vector Machine (SVM) as a trust classifier, which provided the best results compared to Gated Recurrent Unit (GRU) Networks and Extreme Gradient Boosting. The trained SVM was used to predict the user's trust state at each task step and to evaluate the simulated user's trust.

\begin{figure*}[]
\centering
\includegraphics[scale=0.35]{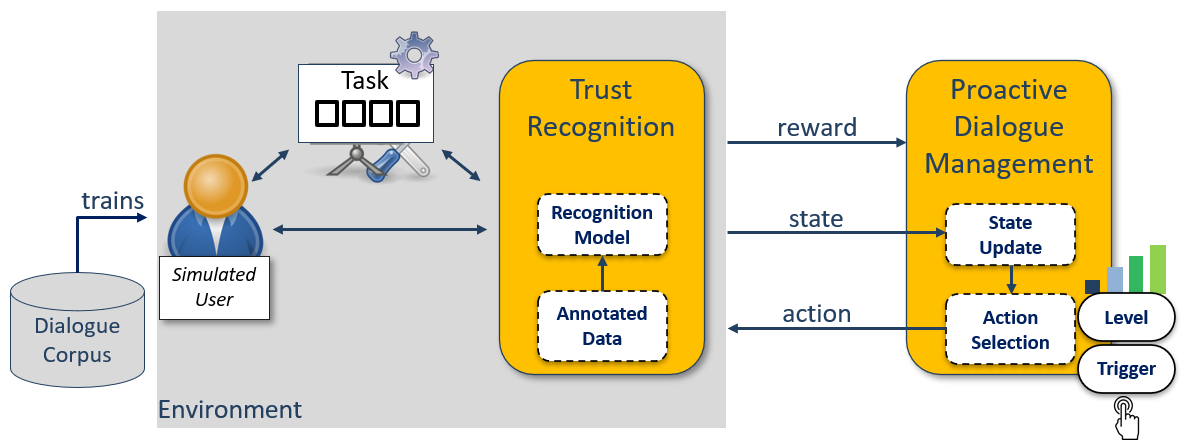}
\caption{The proposed RL framework for implementing trust-adaptive proactive dialog agents. We formulate the collaboration process as an MDP and train an RL-based proactive dialog agent on interactions with a simulated user based on our data collection. For integrating user trust estimations in the state and the reward function of the agent, we utilize our trust estimation module for predicting user trust in the agent's actions in real time.}
\label{img:Architecture}
\end{figure*}

In summary, the created simulated environment consisted of a corpus-based user simulator, a supervised learning-based trust state model which allows for estimating the simulated user's current trust in the agent's action and to be able to adapt the dialog to the estimated user trust in the agent, and an RL-based proactive dialog agent, which is extensively described in our previous work \cite{kraus2022improving}. In Fig \ref{img:Architecture}, the schematic description of the simulated environment is depicted. In the following, we present the architecture of the user simulation model.
\section{User Simulator Architecture}
User simulation was based on two components  similar to the work by \citet{jain2018user}: a \textit{user model} and a \textit{user dialog manager}. The user model contained all the necessary information for modeling distinct user types whose specific task and trust behaviors were imitated. The user dialog manager was designed as a rule-based agent that triggered various behaviors dependent on the proactive system's actions and the current task context. See Table \ref{table:values} for an overview of all parameters used for user simulation.  First, the user model is described in detail.
\subsection{User Model}
\begin{table*}
\small
\centering
\begin{tabularx}{\textwidth}{p{3cm}p{9cm}X}
\hline
\textit{Parameter} & \textit{Description} & \textit{Type}\\
\hline
Age &  numerical value & user trait \\
Gender &   categorical value: male, female, other & user trait\\
Technical Affinity& avg. numerical value based on 5-point Likert scale by~\citet{karrer2009technikaffinitat} & user trait\\
Trust Propensity& avg. numerical value based on 5-point Likert scale by~\citet{merritt2013trust} & user trait\\
Domain Expertise &  avg. numerical value based on self-developed 5-point Likert scale & user trait\\
Big 5 personality traits& avg. numerical value for each trait based on 5-point Likert scale by~\cite{rammstedt2013kurze} & user trait\\
Proactive DialAct & categorical value: None, Notification, Suggestion, Intervention & system action\\
Task Difficulty &  numerical value based on 5-point Likert scale & task property\\
Task Complexity & numerical value: 3,4,5 & task property\\
Task Duration  &  numerical value & task property\\
User Selection &  numerical value according to game score of the respective option & user action\\
Suggestion Request & categorical value: True, False & user action\\
Help Request &  categorical value: True, False & user action\\
\hline
\end{tabularx}
\caption{Overview and description of the user and interaction parameters from the ProDial-corpus.}
\label{table:values}
\end{table*}
To create distinct user types, we utilized user-dependent information from the corpus, including age, gender, technical proficiency, the propensity to trust, domain expertise, and the Big 5 personality traits. We generated random distributions for these variables, with the exception of gender, which was based on its likelihood of occurrence in the corpus. Truncated Gaussian distributions were used for all other variables, as they were rated on 5-point Likert scales or were bounded due to study restrictions (user age was limited from 18 to 60). Our definition of a user's task behavior comprised the selection of options, which was represented as the game score, help requests about the game and suggestion requests towards the system. 
While all user-type features were used for trust estimation, only three variables (domain expertise, propensity to trust, and technical proficiency) were deemed relevant to the specific user's task behavior. Domain expertise influences decision-making and novices may ask for more recommendations. Propensity to trust affects a user's reactions to proactive behavior, with low trust leading to rejections of offers or not asking for assistance. Technical proficiency affects decision-making when collaborating with an autonomous technical system.
To simplify the selection of specific task behavior, these three user traits were transformed into binary values based on a threshold of 3 on the 5-point Likert scale. This was done to reduce the rule space for simulating user behavior.
User-specific task behavior and the system's actions affect task duration and perceived difficulty. Both variables were randomized also using truncated Gaussian distributions as task duration was always greater than 20 seconds and had timing limitation, and perceived difficulty was measured on a 5-point Likert scale. The user's task behavior was based on a pre-defined rule set generated by a user dialog manager, which is described in the next section.
\subsection{User Dialog Manager}
Two distinct approaches, namely complexity-based and task-step-based, were used to generate task behavior in the serious dialog game. The game had 12 task steps, each with a varying level of complexity, which referred to the number of options a user had to select from for decision-making. The number of options per task varied sequentially in the order of 3, 4, 5, 3, 4, and 5. The complexity-based method simulated user behavior based on the system's action and the complexity level of the current task step. For instance, if the current proactive dialog act type was Notification and the current task step had a complexity level of 3, the simulator would use the corpus data distributions for these specific cases for generating task behavior.
On the other hand, the task-step-based method incorporated information from a specific task step and the system's action to simulate user behavior. For instance, if the current proactive dialog act type was Notification and the user was working on the seventh task step, the simulator would use the corpus data distributions for these specific cases for generating task behavior. The advantage of incorporating task complexity in dialog management was that user behavior could be generated in a more generalized way, not dependent on the specific task steps. However, the task-step-based method modeled sequential dependencies between task steps better. Hence, there existed a certain trade-off between both variants.

The simulation of user behavior in both approaches involved generating values for the game score, whether a user-initiated a suggestion or help request, the corresponding duration of the task step, and perceived task difficulty. The probabilities for each specific user behavior were based on structured datasets that depended on the user model and the current dialog situation. The overall dataset was sorted based on the occurrences of user behavior, dependent on the relevant user traits, such as domain experience, propensity to trust, and technical affinity. These user traits were represented as tuples of three binary values, i.e., "000" to "111". For instance, "000" represented low domain experience, low propensity to trust, and low technical affinity, while "111" represented high domain experience, high propensity to trust, and high technical affinity. In the complexity-based method, the user-dependent data was first summarized based on tasks of the same complexity, and then it was summarized according to the types of assistant proactivity, i.e., None, Notification, Suggestion, and Intervention. Lastly, the resulting data was structured according to the occurrences of help and suggestion requests, represented as binary values. In contrast, the task-step-based approach used the same method but summarized the user-dependent data based on the respective task step number. In case the occurrences of specific parameters did not exceed a specific threshold, both approaches used fallback datasets. If there was not enough data for a particular user trait, i.e., occurrences for a trait were below 10, then this parameter was omitted, and the means and standard deviations or counts of all user traits were used for calculating the probabilities for user behavior generation.

The following outlines the process for simulating using the task-step-based approach \footnote{The code for the user simulator is available online at \url{https://github.com/MattKraus90/ProactiveTraining} for reproducibility.}. Similarly, the complexity-based approach algorithm was structured, but with complexity-based data distributions instead of task-step-based distributions. The algorithmic process begins with generating a user type with specific traits and loading approach-specific structured data sets. Then, the dialog game is initialized, and the task steps (1-12) with respective complexities (3, 4, 5) are iterated. For each task step, values for help and suggestion requests, duration, perceived difficulty, and the achieved game score are calculated based on the user type and the system's action. Relevant trait categories are queried, and context is determined, such as proactive system action and complexity or task step number. The fallback threshold is checked, and if not exceeded, personality traits are neglected, and overall means are used for probability calculation. A simulation is conducted to determine if a help and/or suggestion request would be set, with a fallback check for respective request types. Depending on the specific case, the perceived difficulty, task step duration, and the achieved game score are simulated.  
\section{Experiments and Results}
To determine the most suitable approach for training and testing proactive dialog strategies that adapt to trust, we conducted an evaluation that focused on assessing the realism of each user simulation method. Our goal was to identify the approach that best approximated the actual behavior of users, as observed in the data. To achieve this, we employed both methods to simulate user behavior, based on the user types and system actions recorded during data collection with real users. We then compared the simulated behavior with the real behavior, using the KL distance to measure the difference between the two distributions. This approach has been suggested by previous researchers such as \citet{pietquin2013survey} and was used in a study by \citet{jain2018user}. KL distance is a measure of how one probability distribution $Q$ is different from a second, reference probability distribution $P$:
\begin{equation}
D_{KL}(P\parallel Q)=\sum_{x\in \mathcal{X}}P(x)\log \left(\frac{P(x)}{Q(x)}\right)
\end{equation}
where the distances range from 0, i.e. distributions are equal, to 1, i.e distributions are completely different. The lower the distance between the distributions, the more realistic is the respective user simulator. For evaluation, we calculated the distances of distributions between the complexity-, respectively task-step-based approach and actual behavior for each task step. In Table \ref{table:usersim}, the overall mean distances for each task step as well as the individual distances for the game score, duration, help and suggestion request, and perceived difficulty are listed.
\begin{table*}[ht!]
\centering
\small
\begin{tabularx}{\textwidth}{p{2.5cm}|X|X||X|X}
\hline 
& \multicolumn{2}{c}{\textbf{Complexity-based} \textit{M (SD)} } & \multicolumn{2}{c}{\textbf{Task-step-based} \textit{M (SD)} } \\
&  KL & MSE & KL & MSE  \\
\hline \hline
\textbf{Game Score} & 0.369 (.185) & 73.19 (64.6) & 0.354 (.166) & 70.94 (64.7)  \\
\hline
\textbf{Duration} & 0.261 (.064) & 1722 (844) & 0.244 (.079) & 1530 (104)  \\
\hline
\textbf{Difficulty} & 0.145 (.011) & 1.909 (.155 )& 0.149 (.008) & 1.887 (.217) \\
\hline
\textbf{Help Request} & 0.029 (.009) & 0.088 (.028) & 0.031 (.011) & 0.097 (.035) \\
\hline
\textbf{Suggestion Request} & 0.084 (.006) & 0.352 (.025) & 0.082 (.010) & 0.337 (.034)\\
\hline
\hline
\textbf{Overall} & 0.178 (.151) &  359.5 (780) & 0.172 (.142) & 320.6 (765)\\
\hline
\end{tabularx}
\caption{Descriptive statistics of the KL distances and MSEs for each user simulator type with regard to the measures of game score, duration, help and suggestions request, and perceived difficulty.}
\label{table:usersim}
\end{table*}
\section{Discussion and Conclusion}
The study findings indicated that both types of user simulators had similar performance as there were no notable differences in all measured features (with all p-values being greater than 0.05). Table \ref{table:usersim} reveals that the task-step-based approach produced behavior that was slightly more realistic than the complexity-based method. This was mainly attributed to the fact that the former approach simulated more authentic game scores and durations of specific task steps. One possible explanation for this was that the task-step approach used averages of individual task steps to generate distributions, while the complexity-based method used average values across four task steps of the same complexity. Help and suggestion requests were simulated nearly the same way as observed during data collection. The result of the simulators was in the same region as the performance of the user simulator by \citet{jain2018user} by achieving a KL-score of 0.109 which they then used to build an RL-based dialog manager \cite{pecune2020framework}. Thus, we deem both applicable for training and testing statistical proactive dialog strategies. Given the slightly more realistic outcomes, the task-step-based approach was chosen for constructing the train- and test environment in developing statistical proactive dialog. We trained and tested an RL-based proactive dialog agent using the described user simulation approach and presented the results in \citet{kraus2022improving}. The evaluation showed the utility of our approach, as the user simulator provided similar on task effectiveness and user trust as observed in studies with real users \cite{rau2013effects,kraus2020effects}.

As usual, our approach has limitations. Firstly, although we found that 308 participants provided meaningful results for creating user types, increasing the number of participants for data collection could be beneficial. This is especially important for very specific user types where there is limited information, necessitating the need for fallback strategies. Gathering data from more users could alleviate this issue. Secondly, we only focused on a restricted decision-making interaction where users used template natural language utterances to interact with the agent. Therefore, we used a simplistic approach which was deemed reasonable for the task at hand. However, for future work, using more sophisticated approaches such as Hidden Markov Models or RL-based approaches could be beneficial, especially when extending our approach to utilize a greater number of dialog acts. In addition, we aim to integrate natural language interaction, for which transformer-based user simulation approaches could be helpful in creating sophisticated models for more complex use cases. Nonetheless, our approach is among the first to incorporate user, task, and system information for modeling a user's task and trusting behavior during mixed-initiative interaction with a proactive dialog agent. Thus, this work is an important step towards enabling socially responsible and task-effective HAITs.
\bibliographystyle{ACM-Reference-Format}
\bibliography{sample-base}


\end{document}